\documentclass[letterpaper]{article} 
\usepackage[]{aaai25}  
\usepackage{times}  
\usepackage{helvet}  
\usepackage{courier}  
\usepackage[hyphens]{url}  
\usepackage{graphicx} 
\usepackage{natbib}  
\usepackage{caption} 
\usepackage{algorithm}
\usepackage{algorithmic}

\usepackage{newfloat}
\usepackage{listings}
\DeclareCaptionStyle{ruled}{labelfont=normalfont,labelsep=colon,strut=off} 
\floatstyle{ruled}
\newfloat{listing}{tb}{lst}{}
\floatname{listing}{Listing}
\usepackage{xspace}
\usepackage{xcolor}
\usepackage{booktabs} 
\usepackage{colortbl}
\usepackage{svg}
\usepackage{amssymb}
\usepackage{adjustbox}
\usepackage{multirow}
\usepackage{makecell}

\definecolor{wkred}{RGB}{255, 190, 190}
\definecolor{wkblue}{RGB}{210, 230, 250}
\definecolor{wkgold}{RGB}{255, 223, 129}
\definecolor{wksilver}{RGB}{192, 192, 192}
\definecolor{backblue}{RGB}{210, 230, 250}
\definecolor{backred}{RGB}{255, 190, 190}
\definecolor{backgreen}{RGB}{219,242,214} 

\definecolor{codegreen}{rgb}{0,0.6,0}
\definecolor{codegray}{rgb}{0.5,0.5,0.5}
\definecolor{codepurple}{rgb}{0.58,0,0.82}
\definecolor{backcolour}{rgb}{0.95,0.95,0.92}
\definecolor{wkgreen}{RGB}{184,244,175}
\definecolor{wkpurple}{RGB}{210,210,253}
\definecolor{wkyellow}{RGB}{255,241,177}

\newcommand{\method}{{Tangram}\xspace}

\nocopyright 

\title{\method:   Benchmark for Evaluating Geometric Element Recognition in Large Multimodal Models}

\author {
    Chao Zhang\quad\quad Jiamin Tang
    \quad\quad Jing Xiao$^{\thanks{Corresponding author}}$ \\
}
\affiliations{
	South China Normal University \\
	zhangchao@m.scnu.edu.cn \quad\quad xiaojing@scnu.edu.cn
}
\usepackage{bibentry}

\begin{document}

\maketitle

\begin{abstract}
Significant advancements in Large Multimodal Models (LMMs) have enabled them to tackle complex problems involving visual-mathematical reasoning. However, their ability to identify geometric elements remains underexplored. To address this gap, we introduce \method, a novel benchmark designed to evaluate the performance of LMMs on geometric element recognition. \method comprises 1,080 diverse geometric diagrams sourced from primary and secondary school exams, competitions, and textbooks, ranging from simple geometric shapes to complex combinations. Each diagram is paired with four questions, resulting in 4,320 visual-question-answer pairs. Unlike existing benchmarks that emphasize higher-level cognition and reasoning, \method focuses on understanding geometric elements, requiring models to perform a ``simple yet challenging" counting task. Systematic evaluation of 13 prominent LMMs, such as GPT-4o and Claude 3.5 Sonnet, reveals that these models face significant challenges even in seemingly straightforward tasks. The top-performing model achieves an accuracy of only 53.0\%, highlighting a substantial gap compared to human performance. These findings underscore the limitations of current multimodal AI systems in handling basic perception tasks and serve to inspire the development of the next generation of expert-level multimodal foundational models. The data and code will be released soon.
\end{abstract}

%

\section{Introduction}
\begin{figure}[t!]
	\centering
	\includegraphics[width=0.98\columnwidth]{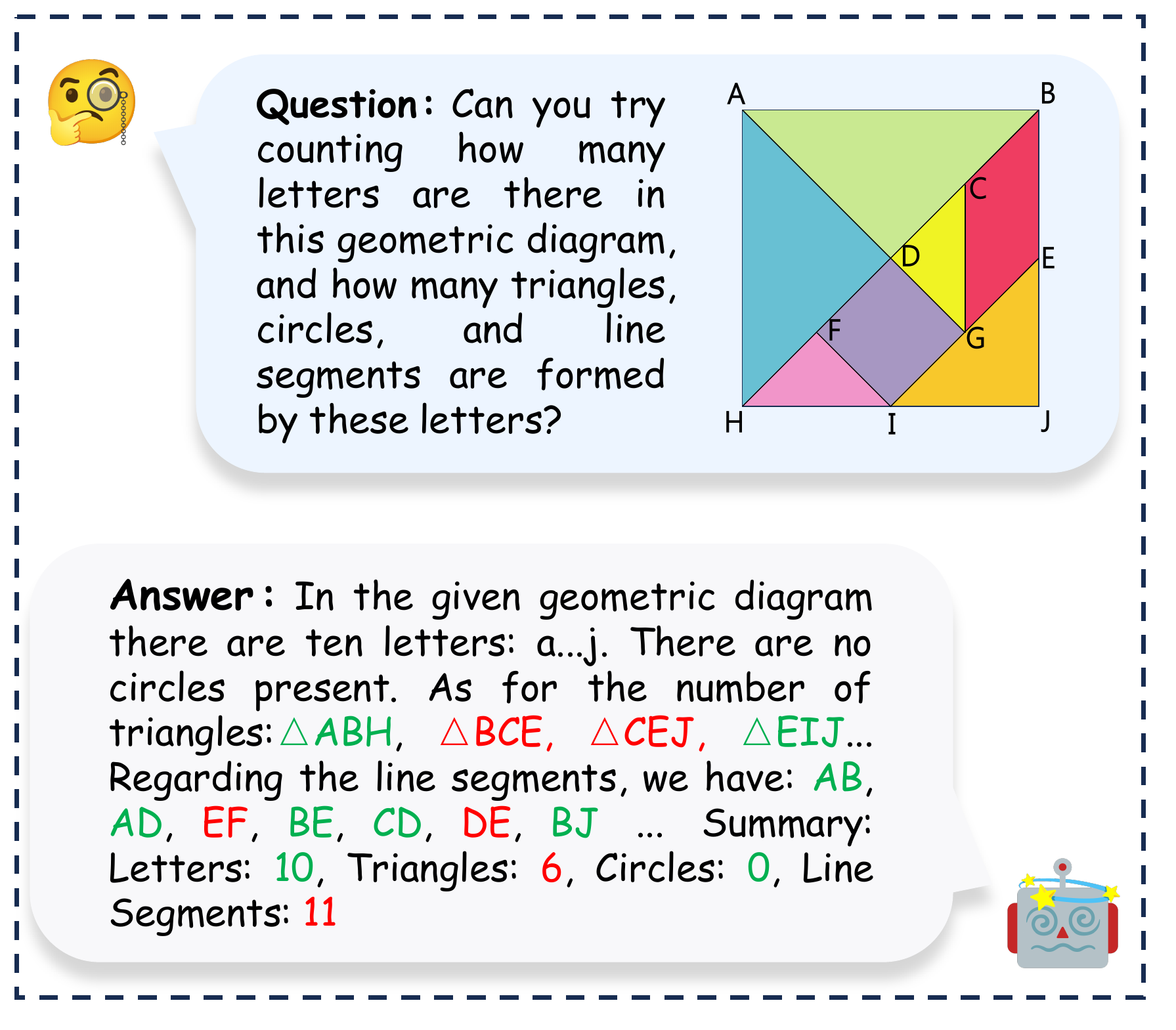}
	\caption{Toy example of testing GPT-4o's accuracy in recognizing geometric elements in the given diagram, with correct answers highlighted in green and errors highlighted in red.}
	\label{fig:wrong-example}
\end{figure}
Recent years have witnessed a technological revolution driven by large language models (LLMs) such as GPT-4o~\cite{GPT-4o} and Claude 3.5 Sonnet~\cite{Claude-3.5-Sonnet}, as well as large multimodal models (LMMs). These models have demonstrated exceptional performance in natural language processing (NLP) and computer vision (CV), igniting widespread academic interest in their potential for tackling higher-level cognitive tasks. In particular, the evaluation of mathematical reasoning capabilities in LLMs and LMMs has become a critical metric in the pursuit of artificial general intelligence (AGI).

Early benchmarks such as SVAMP~\cite{SVAMP}, MATH~\cite{MATH}, and GSM8K~\cite{GSM8K} were commonly used to assess the mathematical reasoning capabilities of large language models (LLMs). As large multimodal models (LMMs) have rapidly developed, researchers have introduced multimodal benchmarks like MathVista~\cite{MathVista}, MATH-Vision~\cite{Math-Vision}, and MathVerse~\cite{MathVerse} to more comprehensively evaluate LMMs' performance on visual context-based mathematical reasoning tasks. The difficulty of these benchmarks has progressively increased, with OlympiadBench~\cite{OlympiadBench} setting a standard comparable to that of Olympic-level mathematics competitions. Despite these advancements, experimental results indicate a significant performance gap between LMMs and humans in multimodal mathematical reasoning. For instance, the authors of MathVerse observed that many LMMs achieve even higher accuracy without visual input, underscoring their challenges in interpreting mathematical diagrams.

Solving geometric problems requires the integration of multimodal information, spatial reasoning, and common sense knowledge, making it a longstanding challenge in artificial intelligence~\cite{GeoQA, Geometry3K, GeoQA+}. The GeoEval~\cite{GeoEval} benchmark, designed to assess the performance of large multimodal models (LMMs) in geometric reasoning, has revealed that while current multimodal foundation models exhibit certain general capabilities, their performance in geometric tasks remains significantly inferior to that of humans. Experiments have shown that LMMs achieve notably higher accuracy when working with pure text-based problems. A series of results highlights the difficulty LMMs face in interpreting mathematical diagrams. To illustrate this, we conducted a simple test on GPT-4o (see Figure \ref{fig:wrong-example}), and its response clearly demonstrated a deficiency in understanding geometric diagrams. However, specialized benchmarks have yet to assess whether LMMs truly comprehend geometric diagrams. Therefore, developing a benchmark tailored specifically to evaluate the geometric diagram comprehension abilities of LMMs is critical for advancing geometric reasoning tasks.

Based on \method, we conducted extensive experiments on several well-known closed-source multimodal models, including GPT-4o, Gemini 1.5 Pro, Claude 3.5 Sonnet, Qwen-VL-Plus~\cite{Qwen-VL}, and Qwen-VL-Max~\cite{Qwen-VL}, as well as open-source large multimodal models such as InternVL2-40B~\cite{InternVL-2}, Phi-3-Vision-128K-Instruct~\cite{phi}, LLaVA-v1.6-Vicuna-13B~\cite{llavav1.6}, and Yi-VL-34B~\cite{Yi-VL}. The experimental results show that the best-performing model achieved only 53.0\% accuracy, which is significantly lower than the 93.6\% accuracy of middle school students and the 99.5\% accuracy of human experts. Notably, model performance did not improve substantially even with the application of the Zero-shot-CoT~\cite{zero-shot-CoT} method. These findings underscore the limitations of current LMMs in recognizing geometric elements, which may contribute to their poor performance in multimodal geometric reasoning tasks.

We believe that \method will advance research in geometric diagram understanding, thereby promoting the development of mathematical reasoning and opening new avenues for creating more powerful LMMs. In summary, our contributions are as follows:

\begin{itemize}
	\item We propose \method, a benchmark comprising 1,080 geometric diagrams and 4,320 questions, all sourced from primary and secondary school exams, competitions, and textbooks. \method is categorized into three difficulty levels to evaluate the capability of LMMs in recognizing geometric elements.
	\item We introduce a novel evaluation method that allows for a fair comparison of various models' performance on \method.
	\item We conduct extensive experiments on multiple existing LMMs, and the results reveal that most models struggle to accurately recognize geometric elements. Even in our designed simple counting tasks, the performance gap between LMMs and humans remains substantial, highlighting significant potential for improvement.
	\item  Benefiting from the fine-grained annotations we provide, we believe \method will offer valuable insights into the future development of LMMs for geometric mathematical reasoning and multimodal reasoning.
\end{itemize}

\begin{figure*}[t!]
	\centering
	\includegraphics[width=1\textwidth]{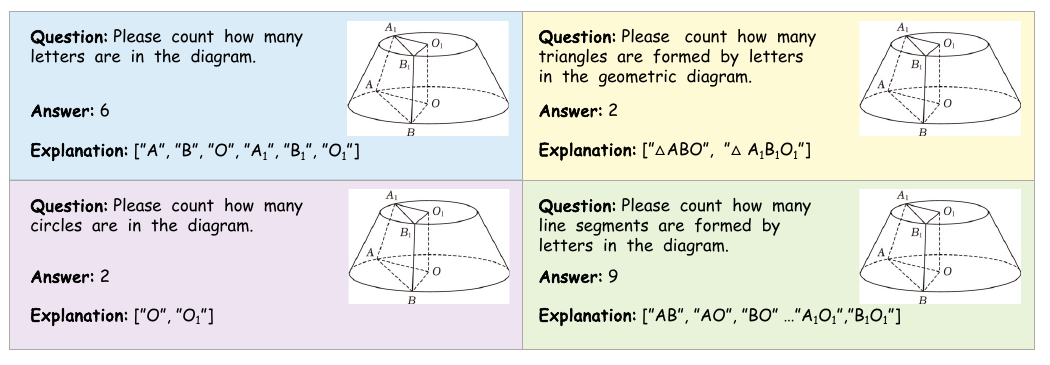}
	\caption{An example from our proposed \method. Each diagram is paired with four questions that involve counting geometric elements, including letters, circles, triangles and line segments.}
	\label{fig:question-example}
\end{figure*}

\section{Related Work}
\subsection{Large Multimodal Models}
Large multimodal models (LMMs) have achieved significant advancements building upon the success of large language models (LLMs) such as GPT~\cite{GPT-1, GPT-2, GPT-3}, LLaMA~\cite{LLaMA-1, LLaMA-2, LLaMA3.1}, and Vicuna~\cite{Vicuna}. These advancements have enabled effective integration across modalities, bridging the gap between visual and textual representations. Closed-source models like GPT-4o~\cite{GPT-4o}, Gemini 1.5 Pro~\cite{Gemini-1.5-pro}, and Claude 3.5 Sonnet~\cite{Claude-3.5-Sonnet} have demonstrated state-of-the-art performance across various benchmarks. Simultaneously, the open-source community has contributed numerous LMMs. For example, LLaVA~\cite{LLaVA, LLaVA-1.5, LLaVA-NeXT} and MiniGPT-4~\cite{MiniGPT-4} employ simple yet effective projection schemes to map image features into the language space. Phi-3-Vision~\cite{phi}, the first multimodal model in the Phi-3 series, integrates text and image processing capabilities, enabling reasoning about real-world images while extracting and inferring textual information. Qwen-VL~\cite{Qwen-VL} enhances multimodal collaboration through a position-aware vision-language adapter. InternVL~\cite{InternVL-1, InternVL-1.5, InternVL-2} introduces a progressive image-text alignment strategy and proposes QLLaMA to bridge the vision encoder with an off-the-shelf LLM decoder. These LMMs have demonstrated impressive performance on benchmarks such as MMMU~\cite{MMMU}, MathVista~\cite{MathVista}, and MathVerse~\cite{MathVerse}. This paper evaluates these models comprehensively using the \method benchmark, specifically assessing their ability to recognize geometric elements.

\subsection{Multimodal Reasoning Benchmarks}
As large multimodal models (LMMs) advance, the academic community has shown increasing interest in evaluating the capabilities. In the early stages, datasets such as NoCaps~\cite{NoCaps}, and Flickr30K~\cite{Flickr30K} are primarily used for image captioning tasks, while VQAv2~\cite{VQAv2}, TextVQA~\cite{TextVQA}, and GQA~\cite{GQA} are applied to visual question answering tasks. Researchers have achieved significant results on these datasets as LMMs have evolved. However, these datasets often focus on single tasks and cannot comprehensively reflect the overall performance of LMMs. Therefore, recent studies have evaluated them from various perspectives. For instance, LVLM-eHub~\cite{LVLM-eHub} collects 47 existing benchmarks to assess six capabilities of LMMs, but it does not create any new benchmarks. While MME~\cite{MME} can comprehensively measure a model's perceptual and cognitive abilities, its question types are relatively simple, requiring only \textit{yes} or \textit{no} answers. Additionally, MMBench~\cite{MMBench} and SEED-Bench~\cite{SEED-Bench} contain a large number of multiple-choice questions, covering a wide range of ability dimensions, but these datasets primarily consist of common-sense questions that do not require extensive domain knowledge or complex reasoning. To strengthen the evaluation of specific domain knowledge, ScienceQA~\cite{ScienceQA} was introduced, covering a wide range of scientific topics from elementary to high school. MathVista~\cite{MathVista}, on the other hand, proposes a series of visually challenging problems, but its scope is limited to the field of mathematics. In comparison, MMMU~\cite{MMMU} includes more difficult expert-level problems that cover 30 different subjects and require nuanced perception, recalling domain-specific knowledge to derive the solution. MathVerse~\cite{MathVerse} focuses on mathematical problems and expands the dataset size to 20K questions. The problems in MATH-Vision~\cite{Math-Vision} are sourced from competitions, but their excessive difficulty hinders the ability to effectively differentiate the performance of various LMMs.

\section{The \method Benchmark}
We introduce \method, a novel benchmark consisting of 1,080 geometric diagrams. As shown in Figure \ref{fig:question-example}, each diagram is accompanied by high-quality annotations detailing the counts of geometric elements within the figure, along with four related questions, resulting in a total of 4,320 visual-question-answer pairs. \method includes both plane and solid geometric diagrams, each containing several geometric elements. The benchmark requires models to count the points, triangles, circles, and line segments in the diagrams, aiming to assess the ability of large multimodal models to recognize geometric elements.

\subsection{Motivation}
\label{sec:motivation}
Large language models (LLMs) and large multimodal models (LMMs) perform well on math word problems; however, their performance is less satisfactory on multimodal math problems, particularly those involving complex reasoning in geometry. Early experiments show that state-of-the-art models achieve accuracy rates of only 54.4\%, 51.0\% and 55.7\% on MathVerse~\cite{MathVerse}, MathVista~\cite{MathVista} and GeoEval~\cite{GeoEval} respectively. These results highlight that large multimodal models still have considerable room for improvement in solving geometry problems. As the old Chinese proverb says, \textit{A journey of a thousand miles begins with a single step}. We argue that accurately understanding the elements within a geometric diagram is a crucial prerequisite for effective reasoning, especially before tackling complex tasks. For geometry problems involving diagrams, the diagrams typically contain rich information, requiring reasoning that integrates both visual and textual elements. It is counterproductive to attempt problem-solving without first identifying the elements within the diagram, as this approach deviates from the natural reasoning process.

\subsection{Data Collection and Annotation}
In the context described above, we develop \method. First, we collect 3,197 geometry problems with corresponding diagrams from online educational websites\footnote{https://www.jyeoo.com/} and textbooks~\cite{book1}. These diagrams are then filtered based on the criterion that each must contain identifiable geometric elements, such as circles and triangles, to ensure fairness and avoid bias in the data. After a rigorous selection process, 1,080 diagrams are included in \method. Finally, the selected diagrams undergo a meticulous annotation phase.

For statistical annotation, we categorize geometric elements into four types: points, circles, triangles, and line segments. We recruit ten master's students majoring in mathematics as annotators to carefully label each diagram in \method. To ensure accuracy and consistency, we implement a double-checking mechanism, where three independent annotators assess each diagram. If their results are inconsistent, the diagram is sent to a specialized review team. This team, composed of experienced senior annotators, re-annotates the disputed diagrams based on detailed guidelines to eliminate uncertainties and ensure the precision of the final annotations. We also design corresponding questions for each category to facilitate a fair evaluation of different large multimodal models in geometric element recognition. An example from \method is shown in Figure \ref{fig:question-example}.

\begin{figure}[t!]
	\centering
	\includegraphics[width=0.95\columnwidth]{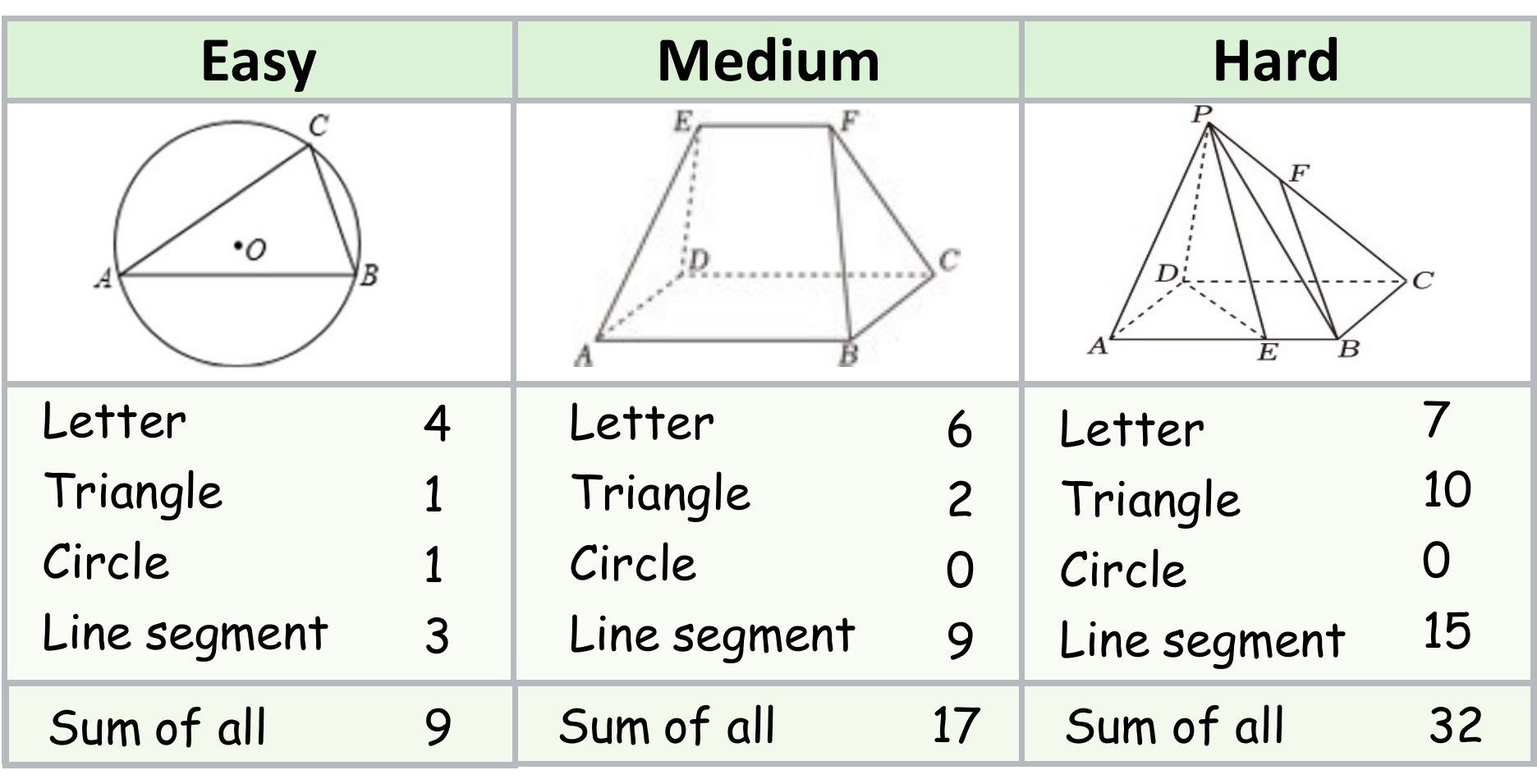}
	\caption{Examples of geometric diagrams in \method categorized by difficulty, showing the number and types of elements in each category.}
	\label{fig:difficulty-example}
\end{figure}

\subsection{Difficulty Classification}
As shown in Figure \ref{fig:difficulty-example}, we categorize the diagrams in \method into three difficulty levels: \textit{Easy}, \textit{Medium} and \textit{Hard}, based on the total number of elements in each diagram. This classification also determines the difficulty of the corresponding questions. For example, when a diagram is labeled as \textit{Hard}, the four associated geometric element counting questions are similarly labeled as \textit{Hard}. Table \ref{tab:difficulty_statistics} provides the statistical details of \method.

\subsection{Features of \method}
To the best of our knowledge, \method is the first benchmark specifically designed to evaluate a model’s ability to recognize geometric elements. Unlike existing benchmarks, \method does not focus on advanced cognition or reasoning, but instead emphasizes tasks that require the model to count geometric elements in diagrams. Below, we further explore the distinctive features of \method.

\subsubsection{Varied Geometric Diagrams \& Elements}
\method includes both solid and plane geometry diagrams, featuring common geometric elements such as points, lines, circles, and triangles. Each diagram consists of multiple independent or overlapping geometric elements.

\subsubsection{Uncontaminated Data}
\method is a novel benchmark with all questions newly constructed as visual question-answer pairs. This approach effectively prevents data leakage, ensuring the fairness of all tests.

\subsubsection{Diverse Challenges}
Each diagram in \method is annotated with a complexity level that corresponds to its associated question. This annotation serves as a guideline for evaluating the proficiency of large multimodal models (LMMs) in recognizing geometric elements.

\setlength{\tabcolsep}{2.3mm}
\begin{table}[t!]
    \centering
    \begin{tabular}{l|ccc}
    \toprule
			\textbf{Sum of elements}& \textbf{Difficulty} & \textbf{Diagram} & \textbf{Question} \\
   \midrule
			0 \textless \ $x$ $\leq$ 15 & Easy & 488  & 1,952\\
			15 \textless \ $x$ $\leq$ 30 & Medium & 459  & 1,836\\
			30 \textless \ $x$ & Hard & 133  & 532\\
			\bottomrule
		\end{tabular}
		\caption{Statistics of our proposed \method, presenting the number of diagrams and questions across three difficulty levels categorized by the sum of elements in each figure.}
		\label{tab:difficulty_statistics}
\end{table}

\setlength{\tabcolsep}{1pt}
\begin{table*}[t!]
	\centering
	\begin{adjustbox}{width=\linewidth}
		\begin{tabular}{l|c|ccccc|ccccc|ccccc}
			\toprule
			\multirow{3}*{\makecell*[l]{\large Model}} &\multirow{3}*{\makecell*[c]{\large All}}
			&\multicolumn{5}{c|}{\makecell*[c]{\shortstack{\method--\textit{Easy}}}} 
			&\multicolumn{5}{c|}{\makecell*[c]{\shortstack{\method--\textit{Medium}}}}
			&\multicolumn{5}{c}{\makecell*[c]{\shortstack{\method--\textit{Hard}}}}\\
			\cmidrule{3-17}
			& &\makecell*[c]{All} &\makecell*[c]{$\S$} &\makecell*[c]{$\equiv$} &\makecell*[c]{$\odot$} &\makecell*[c]{$\bigtriangleup$} &\makecell*[c]{All} &\makecell*[c]{$\S$} &\makecell*[c]{$\equiv$} &\makecell*[c]{$\odot$} &\makecell*[c]{$\bigtriangleup$} &\makecell*[c]{All} &\makecell*[c]{$\S$}&\makecell*[c]{$\equiv$} &\makecell*[c]{$\odot$} &\makecell*[c]{$\bigtriangleup$} \\
			\midrule
			\multicolumn{17}{c}{\textit{Open-source LMMs}}\\
			\cmidrule{1-17}
			LLaVA-v1.6-Vicuna-13B & 20.6& 22.2&26.4&8.2&36.3&18.0&20.8&9.8&1.1&58.4&\colorbox{backblue!50}{13.7}&18.8&6.8&0.0&64.7&3.8 \\
			Phi-3-Vision-128K-Instruct & 22.7& 29.0&23.6&12.1&60.5&19.9&24.3&30.7&1.3&51.9&13.5&14.8&12.8&0.0&43.6&3.0 \\ 
			InternVL2-26B & 23.4& 32.7&36.1&18.6&70.1&6.1&22.5&32.5&5.9&43.4&8.5&14.8&25.6&3.0&26.3&4.5 \\   
			LLaVA-v1.6-Vicuna-7B & 24.2& 25.6&19.3&7.8&52.9&22.5&24.1&7.0&1.7&74.7&12.9&22.9&0.0&0.0&\colorbox{backblue!50}{88.0}&3.8 \\
			Yi-VL-34B & 28.4& 41.2&49.0&14.8&74.6&\colorbox{backblue!50}{26.6}&28.4&30.1&0.0&71.7&12.0&15.6&12.0&0.0&43.6&\colorbox{backblue!50}{6.8} \\
			
			InternVL2-Llama3-76B & 37.2& \colorbox{backblue!50}{50.5}&83.6&\colorbox{backblue!50}{28.1}&81.1&9.0&38.0&64.7&10.2&69.5&7.4&23.3&45.9&3.8&43.6&0.0 \\
			InternVL2-40B & 39.1& 46.5&62.3&26.8&\colorbox{backblue!50}{88.1}&8.6&37.5&46.6&\colorbox{backblue!50}{10.7}&\colorbox{backblue!50}{85.2}&7.2&33.5&42.1&\colorbox{backblue!50}{6.8}&81.2&3.8 \\
			InternVL2-8B &\colorbox{backblue!50} {42.1}& 50.1&\colorbox{backblue!50}{90.8}&21.3&78.3&10.0&\colorbox{backblue!50}{42.4}&\colorbox{backblue!50}{87.6}&5.7&64.1&12.4&\colorbox{backblue!50}{33.6}&\colorbox{backblue!50}{63.2}&3.0&66.2&2.3 \\
			\cmidrule{1-17}
			\multicolumn{17}{c}{\textit{Closed-source LMMs}}\\
			\cmidrule{1-17}
			Qwen-VL-Plus &30.3&36.2&6.4&19.5&\colorbox{backred!50}{95.9}&\colorbox{backred!50}{23.0}&29.4&6.1&0.0&\colorbox{backred!50}{94.1}&\colorbox{backred!50}{17.4}&25.4&6.8&0.0&\colorbox{backred!50}{94.0}&0.8\\
			Qwen-VL-Max &35.7&40.7&40.6&\colorbox{backred!50}{31.3}&88.3&2.7&37.8&43.6&6.1&89.3&12.2&28.8&20.3&3.0&88.7&3.0\\
			GPT-4o & 39.4& 44.7&45.3&29.9&92.2&11.3&39.4&41.4&\colorbox{backred!50}{14.8}&\colorbox{backred!50}{94.1}&7.2&34.0&24.1&\colorbox{backred!50}{13.5}&93.2&5.3 \\
			Gemini 1.5 Pro & 43.6& 53.2&79.7&27.7&92.4&13.1&44.2&69.9&10.7&89.8&6.5&33.3&35.3&3.0&89.5&5.3 \\
			Claude 3.5 Sonnet &\colorbox{backred!50} {47.1}& \colorbox{backred!50}{55.3}&\colorbox{backred!50}{86.7}&28.1&90.8&15.8&\colorbox{backred!50}{47.8}&\colorbox{backred!50}{77.8}&11.3&90.0&12.0&\colorbox{backred!50}{38.2}&\colorbox{backred!50}{52.6}&2.3&90.2&\colorbox{backred!50}{7.5} \\
			\cmidrule{1-17}
			\multicolumn{17}{c}
			{\textit{Closed-source LMMs (Zero-shot-CoT)}} \\
			\cmidrule{1-17}
			Qwen-VL-Plus &29.5&33.7&30.5&9.6&70.1&\colorbox{backgreen!50}{24.4}&27.8&30.3&3.3&65.1&12.3&20.0&15.8&0.0&60.2&3.8 \\
			Qwen-VL-Max &34.2&39.9&46.9&14.1&78.3&20.1&31.7&46.6&4.6&60.8&\colorbox{backgreen!50}{14.8}&24.0&38.1&0.0&56.4&1.5 \\
			Claude 3.5 Sonnet & 40.0& 50.4&86.7&24.0&79.9&11.1&41.3&79.1&10.5&63.0&12.6&28.4&61.7&3.8&42.1&6.0 \\
			Gemini 1.5 Pro & 51.9& \colorbox{backgreen!50}{58.1}&97.1&\colorbox{backgreen!50}{28.1}&92.8&14.5&51.5&94.3&9.6&93.7&8.3&46.1&80.5&3.0&94.0&\colorbox{backgreen!50}{6.8} \\ 
			GPT-4o &\colorbox{backgreen!50}{53.0}& 57.0&\colorbox{backgreen!50}{98.6}&27.5&\colorbox{backgreen!50}{93.4}&8.6&\colorbox{backgreen!50}{53.3}&\colorbox{backgreen!50}{95.6}&\colorbox{backgreen!50}{13.9}&\colorbox{backgreen!50}{95.2}&8.3&\colorbox{backgreen!50}{48.7}&\colorbox{backgreen!50}{88.7}&\colorbox{backgreen!50}{6.0}&\colorbox{backgreen!50}{97.0}&3.0 \\
			\cmidrule{1-17}
			\multicolumn{17}{c}
			{\textit{Human performance}} \\
			\cmidrule{1-17} 
			\textit{Human-student} &93.6&96.5&98.7&94.8&100&92.4&93.8&96.4&90.5&100&88.3&90.7&94.2&85.6&100&98.1\\
			\textit{Human-expert} &99.5&99.9&100&99.8&100&99.7&99.6&100&99.5&100&98.9&99.1&99.9&98.3&100&98.1
			\\
			\bottomrule
		\end{tabular}
	\end{adjustbox}
	\label{supp-t4}
	
	\caption{\textbf{Accuracy(\%) scores of models on our \method.} ALL: overall accuracy. {$\S$}: Letter; {$\equiv$}: Line Segment; {$\odot$}: Circle; {$\bigtriangleup$}: Triangle. The highest accuracy for {open-source}, {closed-source} and closed-source{(Zero-shot-CoT)} LMMs is marked in \colorbox{backblue!75}{blue}, \colorbox{backred!50}{red} and \colorbox{backgreen!50}{green} respectively.}
	\label{tab:main_results}
\end{table*}

\section{Experiments}
In this section, we present systematic experiments on \method, evaluating both open-source and closed-source large multimodal models (LMMs). Our results show that, even for simple counting tasks that humans can easily perform, the accuracy of the most advanced LMMs remains low. Moreover, we compare the performance of open-source and closed-source models, demonstrating that closed-source models generally exhibit superior overall performance. We also analyze the models' recognition capabilities across different geometric elements and provide detailed insights into these findings.

\subsection{Evaluation Metric}
We use accuracy as the evaluation metric to fairly compare the performance of various models on the \method benchmark. Following the approach outlined in MathVista~\cite{MathVista}, we extract answers from the models' responses. Specifically, we combine the answer extraction prompt\footnote{Details on prompt design are provided in the Appendix.} with the model-generated response into a single sequence, which is then input into GPT-4o~\cite{GPT-4o} to extract the target value. By manually comparing the extracted answers with the ground truth across 400 sampled examples, we found an accuracy exceeding 98\%, similar to the success rate reported in MathVista~\cite{MathVista}. Finally, we compare the extracted results with the benchmark’s standard answers to calculate the final accuracy.

\subsection{Experimental Setup}
\subsubsection{LMMs}
We evaluate two types of foundational models, open-source and closed-source models on \method. Closed-source models are assessed using their official APIs, while open-source models are evaluated by running inferences on 8 NVIDIA A100 40GB GPUs. For closed-source models, we select the most representative ones, including GPT-4o~\cite{GPT-4o}, Gemini 1.5 Pro~\cite{Gemini-1.5-pro}, Claude 3.5 Sonnet~\cite{Claude-3.5-Sonnet}, Qwen-VL-Plus~\cite{Qwen-VL}, and Qwen-VL-Max~\cite{Qwen-VL}. For open-source models, we consider a range of model sizes, from 7B to 76B, including LLaVA-v1.6-Vicuna-7B~\cite{llavav1.6}, Yi-VL-34B~\cite{Yi-VL}, and InternVL2-Llama3-76B~\cite{InternVL-2}. We adopt a zero-shot setting to infer \method questions across all LMMs. Additionally, for closed-source models, we evaluate performance under the Zero-shot-CoT~\cite{zero-shot-CoT} setting. The prompts and hyperparameters used for all LMMs are provided in the Appendix. For open-source LMMs, we conduct experiments using LMDeploy\footnote{https://github.com/InternLM/lmdeploy/}.

\subsubsection{Human}
We recruited ten middle school students, each of whom completed element-counting tasks on 108 geometric diagrams. The results of these participants are labeled as \textit{Human-student} in our experimental results. Additionally, three graduate students majoring in mathematics were recruited to complete the tasks, and their results are labeled as \textit{Human-expert}.

\subsection{Experimental Results}
Table \ref{tab:main_results} presents the overall experimental results. Based on these results, our key findings can be summarized as follows:

\paragraph{Challenging nature of \method}
Table \ref{tab:main_results} demonstrates the challenges posed by \method for current Large Multimodal Models (LMMs). While GPT-4o achieves the best performance with an accuracy of 53.0\%, there remains a significant gap compared to \textit{Human-student} (93.6\%) and an even larger gap compared to \textit{Human-expert} (99.5\%).

\subsubsection{Closed-source LMMs are better-performed}
\begin{figure}
	\centering
	\includegraphics[width=1\columnwidth]{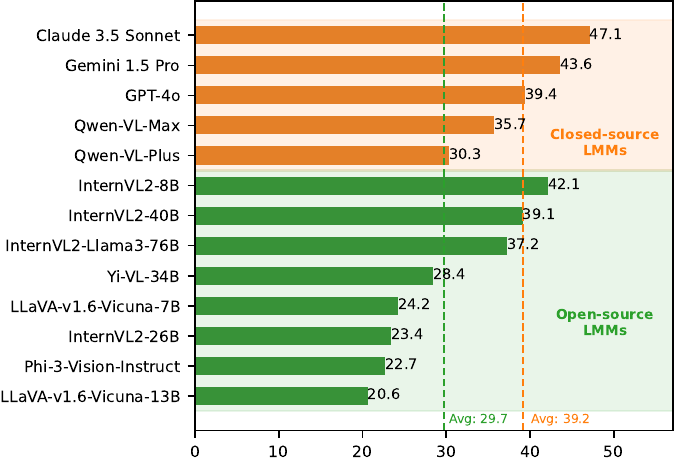}
	\caption{Accuracy(\%) comparison between closed-source and open-source LMMs.}
	\label{fig:compare_model}
\end{figure}

According to the experimental results, closed-source models generally outperform open-source models. Notably, InternVL-8B achieves an accuracy of 42.1\%, approaching the 43.6\% accuracy of Gemini 1.5 Pro. Additionally, InternVL-8B outperforms closed-source models such as Qwen-VL-Plus, Qwen-VL-Max, and even GPT-4o. These findings suggest that while open-source models have made substantial progress, there remains significant room for improvement. Furthermore, as shown in Figure \ref{fig:compare_model}, closed-source models exhibit an average accuracy of 39.2\%, which is 9.5\% higher than the 29.7\% average accuracy of open-source models.

\subsubsection{Comparison among different geometric elements}
\begin{figure}[t!]
	\centering
	\includegraphics[width=1\columnwidth]{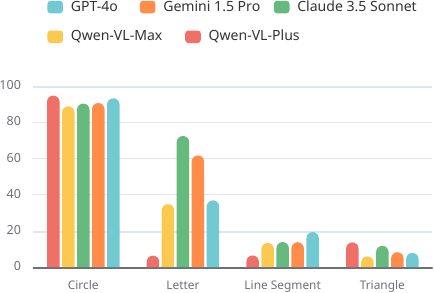}
	\caption{Performance of models on different types of geometric elements.}
	\label{fig:performence_on_elements}
\end{figure}

Each diagram in \method contains various geometric elements, and LMMs exhibit significant variation in their ability to recognize these elements. As shown in Figure \ref{fig:performence_on_elements}, the recognition accuracy for line segments and triangles is consistently lower than for circles and letters. As the complexity of the diagram increases, the accuracy for line segments and triangles drops sharply. We speculate that this is due to the frequent overlap of triangles and lines within a single diagram, which becomes more pronounced as the diagram’s complexity rises. This finding suggests that current LMMs struggle with recognizing overlapping elements. Furthermore, as shown in Table \ref{tab:plane-solid-diagram}, model accuracy on solid geometric diagrams is significantly lower than on plane diagrams. This may be because solid geometry is generally more complex and abstract, making it more challenging for models to interpret.

\setlength{\tabcolsep}{10pt}
\begin{table}[t!]
	\centering
	\begin{tabular}{@{}lcc@{}}
		\toprule
		Model       & \multicolumn{1}{c}{Plane} & \multicolumn{1}{c}{Solid} \\ \midrule
		GPT-4o        & 55.7                & 50.4 (\textit{-5.3})               \\
		Gemini 1.5 Pro & 52.9   & 41.9   (\textit{-11.0})     \\ 
		Claude 3.5 Sonnet      & 51.5      & 46.1 (\textit{-5.4})\\
		Qwen-VL-Max & 30.8  & 26.5    (\textit{-4.3})      \\
		Qwen-VL-Plus        & 27.4                & 23.3    (\textit{-4.1})          \\
		\bottomrule
	\end{tabular}
	\caption{Recognition accuracy(\%) of closed-source LMMs on plane and solid diagrams.}
	\label{tab:plane-solid-diagram}
\end{table}

\subsubsection{Chain-of-Thought prompting is useful but limited}
Following the chain-of-thought (CoT)~\cite{zero-shot-CoT} template, we append ``Let's think step by step" to the original prompt. The detailed Zero-shot-CoT template is provided in the Appendix (Table \ref{tab:instruction-prompt}). As shown in Table \ref{tab:main_results}, using CoT prompting results in an improvement for most models, indicating that CoT facilitates more structured and thorough reasoning. Notably, these improvements are most pronounced in tasks involving simpler geometric elements, such as recognizing letters and circles, where models like GPT-4o and Claude 3.5 Sonnet show significant progress.

However, this benefit is not uniform across all categories. For more complex shapes, such as line segments and triangles, CoT prompting actually reduces recognition accuracy. For instance, in the \method-\textit{Easy} subset, GPT-4o's overall accuracy for letters increases to 98.6\%, but the accuracy for line segments drops from 37.5\% to 24.5\% under the same model. Similarly, CoT prompting leads to a decline in triangle recognition accuracy with Claude 3.5 Sonnet. Our manual checks revealed that when models follow CoT prompts and reason step by step, minor errors in the early stages of reasoning are often amplified in subsequent steps, resulting in increased error propagation.

\subsubsection{Impact of model size on performance}
\begin{figure}[t!]
    \centering
    \includegraphics[width=0.85\columnwidth]{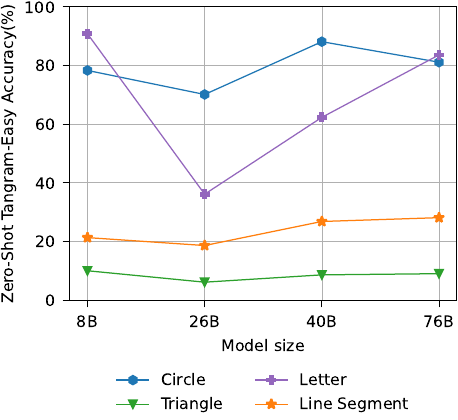}
    \caption{Impact of model size on zero-shot accuracy(\%) on \method-\textit{Easy} for four geometric elements.}
    \label{fig:model_size_performence}
\end{figure}

We conducted experiments on \method using the InternVL2 model family to evaluate the impact of model size on performance. As shown in Figure \ref{fig:model_size_performence}, performance generally improves with increasing model size, though the degree of improvement varies across different types of geometric elements. For example, recognition accuracy for letters significantly improves as model size increases from 26B to 76B, rising from 36.1\% (26B) to 83.6\% (76B). This suggests that larger models are better suited for handling simpler, more structured geometric elements.

In contrast, the improvement is less pronounced for more complex elements such as line segments and triangles, and in some cases, the trend is even inverted. This indicates that increasing model size alone does not necessarily enhance performance on more challenging geometric shapes. The results suggest that the intrinsic complexity of these elements in \method may require advancements in model architecture or training strategies to address these more difficult cases effectively.

\section{Conclusion}
In this paper, we introduce \method, a benchmark designed to assess the ability of large multimodal models (LMMs) to recognize geometric elements. The benchmark consists of 1,080 geometric diagrams and 4,320 questions, with each diagram categorized into one of three difficulty levels. Experiments with both open-source and closed-source models reveal a consistent decline in recognition accuracy as diagram complexity increases. Moreover, a significant performance gap remains between LMMs and human annotators on \method, highlighting the need for further advancements in LMMs' visual comprehension.
\method also underscores that most existing LMMs struggle with accurately recognizing geometric diagrams, particularly when overlapping elements are involved. We hope this work will provide valuable insights into multimodal mathematical reasoning and contribute to improving the visual comprehension capabilities of LMMs.



\bibliography{aaai25}

\clearpage
\appendix
\section{Distributions of The \method}
\begin{figure}[h]
	\centering
	\includegraphics[width=0.75\columnwidth]{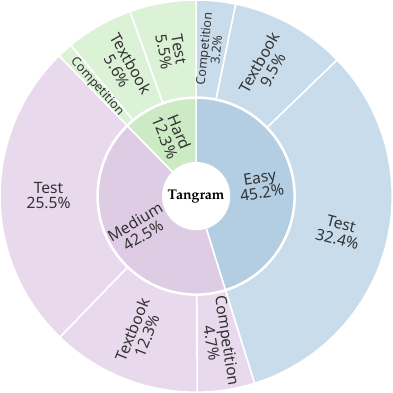}
	\caption{Distributions of the \method benchmark.}
	\label{fig:geoeval-hard-example}
\end{figure}

Figure \ref{fig:geoeval-hard-example} presents a distributions of the \method benchmark. This benchmark consists of a total of 1,080 geometric diagrams, categorized into three difficulty levels. These diagrams are sourced from real-world primary and secondary school exams, competitions, and textbooks.

\section{Evaluation Details}

\subsection{Instruction Prompt Used for Evaluating Models}
\label{appendix:instruction-prompt}
As shown in Table \ref{tab:instruction-prompt}, we combined the question with the diagram and fed them into LMMs to to generate responses. Furthermore, we conducted evaluations using both Zero-shot and Zero-shot-CoT methods. The latter builds upon the former by adding the prompt "Let's think step by step." to guide the model towards stepwise reasoning.

\subsection{Prompt for Answer Extraction}
The prompt used to instruct GPT-4o for answer extraction is illustrated in Table \ref{tab:promt_answer_extraction}.

\subsection{Model Hyper-parameters}
Table \ref{tab:model-hyperparameters} presents the complete list of hyper-parameters applied to the models throughout the evaluation phase. Additionally, Table \ref{supp-t0.5} presents the release time and model sources of LMMs used in Tangram.

\section{Case Study}
Figure \ref{fig:question-example} illustrates the successes and failures of the Qwen-VL-Max model in different tasks. It can be seen that the model performs quite well in recognizing circles and letters. However, when confronted with triangles and line segments that contain a large number of overlapping elements, the model encounters significant difficulties.

\begin{table*}[h!]
	\small
	\centering

	\begin{tabular}{l}
		\toprule
		\textbf{Problem texts} \\ 
		\hline
		Please count how many letters are in the diagram. \\ 
		Please count how many circles are in the diagram. \\ 
		Please count how many triangles are formed by letters in the diagram. \\ 
		Please count how many line segments are formed by letters in the diagram. \\ 
		\bottomrule
	\end{tabular}
	\caption{Four counting task questions designed for LMMs}
	\label{tab:four-questions}
\end{table*}

\begin{table*}[t!]
	\small
	\centering
	\begin{scriptsize}
		\begin{tabular}{@{}lll@{}}
			\toprule
			& \textbf{Template} & \textbf{Example} \\
			\midrule
			Zero-shot &
			\begin{tabular}[c]{@{}l@{}}\$\{problems texts\},\\
				Hint: Please provide the final value, e.g., 1, 2, 3, at the end. \\
				\$\{diagram\}.
			\end{tabular} &
			\begin{tabular}[c]{@{}l@{}}Please count how many line segments are formed by letters in the diagram. \\
				Hint: Please provide the final value, e.g., 1, 2, 3, at the end. \\
				diagram1.png
			\end{tabular}
			\\
			\midrule 
			Zero-shot-CoT &
			\begin{tabular}[c]{@{}l@{}}\$\{problems texts\}, \\
				Hint: Please provide the final value, e.g., 1, 2, 3, at the end. \\
				Let's think step by step.\\ 
				\$\{diagram\},
			\end{tabular} &
			\begin{tabular}[c]{@{}l@{}}Please count how many line segments are formed by letters in the diagram. \\
				Hint: Please provide the final value, e.g., 1, 2, 3, at the end. \\
				Let's think step by step. \\
				diagram2.png
			\end{tabular} \\
			\bottomrule
		\end{tabular}
	\end{scriptsize}
	\caption{Input prompts of LMMs for response generation. The placeholder ``\$\{problems texts\}" represents the four problems(As shown in table~\ref{tab:four-questions}) that ask LMMs to count the four kinds of elements in a geometric diagram. The placeholder ``\$\{diagram\}" represents the corresponding geometric diagram.}
	\label{tab:instruction-prompt}
\end{table*}

\begin{table*}[tbhp!]
	\centering
	\small
	\renewcommand\tabcolsep{3.0pt} 
	\begin{tabular}{cp{12.9cm}}
		\toprule
		\textbf{Element} & \textbf{Prompt} \\
		\midrule
		Task description & 
		\begin{minipage}[s][0.6cm]{1.5\columnwidth}
			You are a result extraction bot.Please read the following example. Then extract the answer from the model response and type it at the end of the prompt.
		\end{minipage}
		\\
		\midrule
		Example 1 & 
		\begin{minipage}[s][1.8cm]{1.5\columnwidth}
			\textbf{Hint:} Please answer the question requiring an integer answer and provide the final value,e.g., 1, 2, 3, at the end.\\
			\textbf{Question:} Can you try counting how many letters are in this geometric diagram?\\
			\textbf{Model response:} There are nine letters (A to I) in the given geometric diagram.
			\\
			\textbf{Extracted answer:} \textbf{9}\\
		\end{minipage}
		\\
		\midrule
		Example 2 & 
		\begin{minipage}[s][2.1cm]{1.5\columnwidth}
			\textbf{Hint:} Please answer the question requiring an integer answer and provide the final value, e.g., 1, 2, 3, at the end.\\
			\textbf{Question:} Can you try counting how many triangles are formed by letters in this geometric diagram?\\
			\textbf{Model response:} The letters form two triangles: Triangle ABD with vertices A, B, D. Triangle BCE with vertices B, C, E.
			\\
			\textbf{Extracted answer:} \textbf{2}
		\end{minipage}
		\\
		\midrule
		Example 3 & 
		\begin{minipage}[s][1.8cm]{1.5\columnwidth}
			\textbf{Hint:} Please answer the question requiring an integer answer and provide the final value, e.g., 1, 2, 3, at the end.\\
			\textbf{Question:} Can you try counting how many circles are formed by letters in this geometric diagram?\\
			\textbf{Model response:} There is only one circle in the image.\\
			\textbf{Extracted answer:} \textbf{1}
		\end{minipage}
		\\
		\midrule
		Example 4 & 
		\begin{minipage}[s][2.8cm]{1.5\columnwidth}
			\textbf{Hint:} Please answer the question requiring an integer answer and provide the final value, e.g., 1, 2, 3, at the end.\\
			\textbf{Question:} Can you try counting how many line segments are formed by letters in this geometric diagram?\\
			\textbf{Model response:} We can count three line segments from the letter points: Line segment AB between point A and point B. Line segment BC between point B and point C. Line segment CD between point C and point D.\\
			\textbf{Extracted answer:} \textbf{3}
		\end{minipage}
		\\
		\bottomrule
	\end{tabular} 
	\caption{Task-specific instructions for extracting the answer. The table shows three examples with answers that can be extracted using the prompt.}
	\label{tab:promt_answer_extraction}
\end{table*}

\clearpage
\begin{table*}[tbhp!]
	\centering
	\begin{scriptsize}
		\begin{tabular}{@{}lll@{}}
			\toprule
			\multicolumn{1}{c}{Model Name} & \multicolumn{1}{c}{Generation Parameters}                & \multicolumn{1}{c}{Comments}     \\ \midrule
			GPT-4o                  & temperature = 0.0, max\_tokens = 1024 & version = ``gpt-4o-2024-05-13" \\ \midrule
			Gemini 1.5 Pro            & temperature = 0.0, top\_p = 1, max\_tokens = 1024              & version = ``gemini-1.5-pro-001"                                    \\ \midrule
			Claude 3.5 Sonnet         & temperature = 0.0, max\_tokens = 1024                         & version = ``claude-3-5- sonnet-20240620"     \\ \midrule
			Qwen-VL-Max                          & temperature = 0.0, max\_tokens = 1024                        & version = ``qwen-vl-max"    
			\\ \midrule
			Qwen-VL-Plus                & temperature = 0.0, max\_tokens = 1024              & version = ``qwen-vl-plus"
			\\ 
			\midrule
			Yi-VL-34B  &do\_sample = True, temperature = 0.6, top\_p = 0.8, max\_token = 1024    & model =``01-ai/Yi-VL-34B"\\
			\midrule
			Phi-3-Vision-128K-Insstruct  &do\_sample = True, temperature = 0.6, top\_p = 0.8, max\_token = 1024    & model =``microsoft/Phi-3-vision-128k-instruct"\\
			\midrule
			InternVL2-8B & temperature = 0.1, do\_sample = True,  top\_p = 1.0, top\_k = 50, max\_new\_token = 1024 &model = ``OpenGVLab/InternVL2-8B"\\
			\midrule
			InternVL2-26B & temperature = 0.1, do\_sample = True,  top\_p = 1.0, top\_k = 50, max\_new\_token = 1024 &model = ``OpenGVLab/InternVL2-26B"\\
			\midrule
			InternVL2-40 & temperature = 0.1, do\_sample = True,  top\_p = 1.0, top\_k = 50, max\_new\_token = 1024 &model = ``OpenGVLab/InternVL2-40B"\\
			\midrule
			InternVL2-LLama3-76B & temperature = 0.1, do\_sample = True,  top\_p = 1.0, top\_k = 50, max\_new\_token = 1024 &model = ``OpenGVLab/InternVL2-Llama3-76B"\\
			\midrule
			LLaVA-v1.6-Vicuna-7B         & max\_length = 1024                  & model =``liuhaotian/llava-v1.6-vicuna-7b"             \\ 
			\midrule
			LLaVA-v1.6-Vicuna-13B         & max\_length = 1024                  & model =``liuhaotian/llava-v1.6-vicuna-13b"            
			\\ \bottomrule
		\end{tabular}
		\caption{The hyper-parameters for the models used in the evaluation are detailed. When the ``comments" section includes the format \textit{pipeline = ``~"}, it signifies that the model was loaded by the LMDeploy toolkit, where more details can be found in \url{https://github.com/InternLM/lmdeploy}.}
		\label{tab:model-hyperparameters}
	\end{scriptsize}
\end{table*}

\begin{table*}
\centering
        \begin{tabular}{l@{\hspace{0.5cm}}c@{\hspace{2cm}}p{0.5\textwidth}}
            \toprule
            \textbf{Model} & \textbf{\makecell{Release\\ Time}} & \textbf{\makecell[c]{Source}} \\
            \midrule
            Claude 3.5 Sonnet      &   2024-06   & \url{https://claude.ai/} (claude-3-5-sonnet-20241022) \\
            \midrule
            GPT-4o     &   2024-05    & \url{https://platform.openai.com/} (gpt-4o-2024-11-20)\\
            \midrule
            Gemini 1.5 Pro    & 2024-05    & \url{https://ai.google.dev/} (gemini-1.5-pro-002 2024-09-24)\\
            \midrule
            \multirow{2}{*}{Qwen-VL-Max}            & \multirow{2}{*}{2024-01} & \url{https://help.aliyun.com/zh/dashscope/developer-reference/vl-plus-quick-start} \\
            \midrule
            \multirow{2}{*}{Qwen-VL-Plus}        & \multirow{2}{*}{2023-11}  & \url{https://help.aliyun.com/zh/dashscope/developer-reference/vl-plus-quick-start} \\
            \midrule
            \multirow{1}{*}{Yi-VL-34B}        &    \multirow{1}{*}{2024-01}    & \url{https://www.aimodels.fyi/creators/huggingFace/01-ai} \\
            \midrule
            \multirow{1}{*}{Phi-3-Vision-128K-Insstruct}        &    \multirow{1}{*}{2024-05}    & \url{https://huggingface.co/microsoft/Phi-3-vision-128k-instruct} \\
            \midrule
            \multirow{1}{*}{LLaVA-v1.6-Vicuna-7B}       &   \multirow{1}{*}{2023-12}    &        \url{https://huggingface.co/liuhaotian/llava-v1.6-vicuna-7b} \\
            \midrule
            \multirow{1}{*}{LLaVA-v1.6-Vicuna-13B}  &  \multirow{1}{*}{2023-12} & \url{https://huggingface.co/liuhaotian/llava-v1.6-vicuna-13b} \\
            \midrule
            \multirow{1}{*}{InternVL2-8B}       &   \multirow{1}{*}{2024-07}    &        \url{https://huggingface.co/OpenGVLab/InternVL2-8B} \\
            \midrule
            \multirow{1}{*}{InternVL2-26B}       &   \multirow{1}{*}{2024-07}    &        \url{https://huggingface.co/OpenGVLab/InternVL2-26B} \\
            \midrule
            \multirow{1}{*}{InternVL2-40B}       &   \multirow{1}{*}{2024-07}    &        \url{https://huggingface.co/OpenGVLab/InternVL2-40B} \\
            \midrule
            \multirow{1}{*}{InternVL2-LLama3-76B}       &   \multirow{1}{*}{2024-07}    &        \url{https://huggingface.co/OpenGVLab/InternVL2-LLama3-76B} \\
            \bottomrule        
        \end{tabular}
		\caption{{The release time and model source of LMMs used in \method.}}
	    \label{supp-t0.5}
\end{table*}

\begin{figure*}[t!]
    \centering
    \includegraphics[width=1\textwidth]{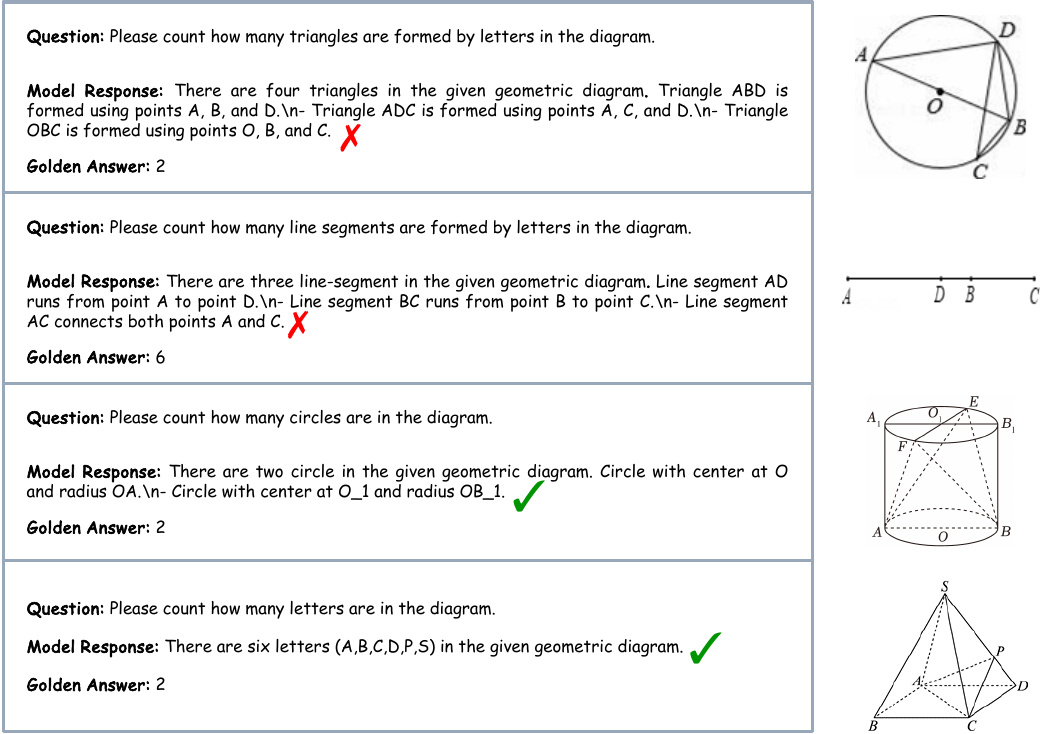}
    \caption{Several successful and failed examples from Qwen-VL-Max.}
    \label{fig:error-question-example}
\end{figure*}

\end{document}